\documentclass[10pt,twocolumn,letterpaper]{article}

\usepackage{cvpr}              

\usepackage{multirow}
\usepackage{array}
\usepackage{algorithm}
\usepackage{algpseudocode}
\usepackage[shortlabels]{enumitem}
\usepackage{stfloats}

\newcommand{\corrauthor}{\textsuperscript{\textdagger}}

\definecolor{cvprblue}{rgb}{0.21,0.49,0.74}
\usepackage[pagebackref,breaklinks,colorlinks,allcolors=cvprblue]{hyperref}

\def\paperID{xxxx} 
\def\confName{CVPR}
\def\confYear{2026}

\title{OnlineX: Unified Online 3D Reconstruction and Understanding \\
with Active-to-Stable State Evolution}

\begin{document}

\vspace{-3mm}
\author{
Chong Xia \quad
Fangfu Liu \quad
Yule Wang \quad
Yize Pang  \quad
Yueqi Duan\corrauthor\\
Tsinghua University \\
Project Page: \url{https://xiac20.github.io/OnlineX/}
}



\maketitle

\begingroup
\renewcommand\thefootnote{} 
\footnote{\corrauthor Corresponding author.}
\endgroup

\begin{abstract}
Recent advances in generalizable 3D Gaussian Splatting (3DGS) have enabled rapid 3D scene reconstruction within seconds, eliminating the need for per-scene optimization. However, existing methods primarily  follow an offline reconstruction paradigm, lacking the capacity for continuous reconstruction, which limits their applicability to online scenarios such as robotics and VR/AR. In this paper, we introduce OnlineX, a feed-forward framework that reconstructs both 3D visual appearance and language fields in an online manner using only streaming images. A key challenge in online formulation is the cumulative drift issue, which is rooted in the fundamental conflict between two opposing roles of the memory state: an active role that constantly refreshes to capture high-frequency local geometry, and a stable role that conservatively accumulates and preserves the long-term global structure. To address this, we introduce a decoupled active-to-stable state evolution paradigm. Our framework decouples the memory state into a dedicated active state and a persistent stable state, and then cohesively fuses the information from the former into the latter to achieve both fidelity and stability. Moreover, we jointly model visual appearance and language fields and incorporate an implicit Gaussian fusion module to enhance reconstruction quality. Experiments on mainstream datasets demonstrate that our method consistently outperforms prior work in novel view synthesis and semantic understanding, showcasing robust performance across input sequences of varying lengths with real-time inference speed.

\end{abstract}
    
\section{Introduction}
3D Gaussian Splatting (3DGS) has recently emerged as a promising alternative for real-time 3D scene reconstruction, offering explicit and efficient representations by rasterizing textured Gaussians~\cite{kerbl20233d, yu2024mip, ren2024octree, luiten2024dynamic}. To eliminate the need for per-scene optimization, generalizable feed-forward models~\cite{chen2024mvsplat, charatan2024pixelsplat, zou2024triplane, zheng2024gps, wewer2024latentsplat, li2024ggrt} have been proposed to directly predict Gaussians from images with known camera poses. However, this reliance on pre-computed poses from offline SfM tools like COLMAP~\cite{schonberger2016structure} has motivated the development of recent pose-free methods~\cite{wang2024dust3r, leroy2024grounding, ye2024no, zhang2025flare} that jointly estimate both poses and scenes.

Despite these advances, most existing approaches follow an offline reconstruction paradigm, making them incompatible with online applications such as robotics, AR/VR, or mobile scanning, where RGB images arrive sequentially, and reconstruction needs to be performed concurrently. Recent works have started to address this online setting. Methods like Spann3R~\cite{wang20243d} and LONG3R~\cite{chen2025long3r} utilize an explicit spatial memory of past frames to assist the current frame prediction, but this leads to significant memory overhead as the sequence grows. 

In contrast, CUT3R~\cite{wang2025continuous} employs a learnable hidden state to store historical information. While the architecture is simple and memory-efficient, it is susceptible to long-term drift, a problem stemming from the representational bottleneck of its single state. Ideally, this hidden state is expected to not only incorporate detailed geometry and appearance cues from neighboring views but also maintain the accurate global structure cues accumulated from all preceding frames. However, as high-frequency local geometry is continuously updated under dense supervision with each new frame, the global information from preceding frames is progressively forgotten, resulting in a drift in the overall structure. Therefore, the key challenge in online reconstruction is reconciling the need to actively integrate local geometry from new observations with the need for a stable, persistent state to ensure long-term global consistency.

\begin{figure*}[t]
    \centering
    \includegraphics[width=0.95\linewidth]{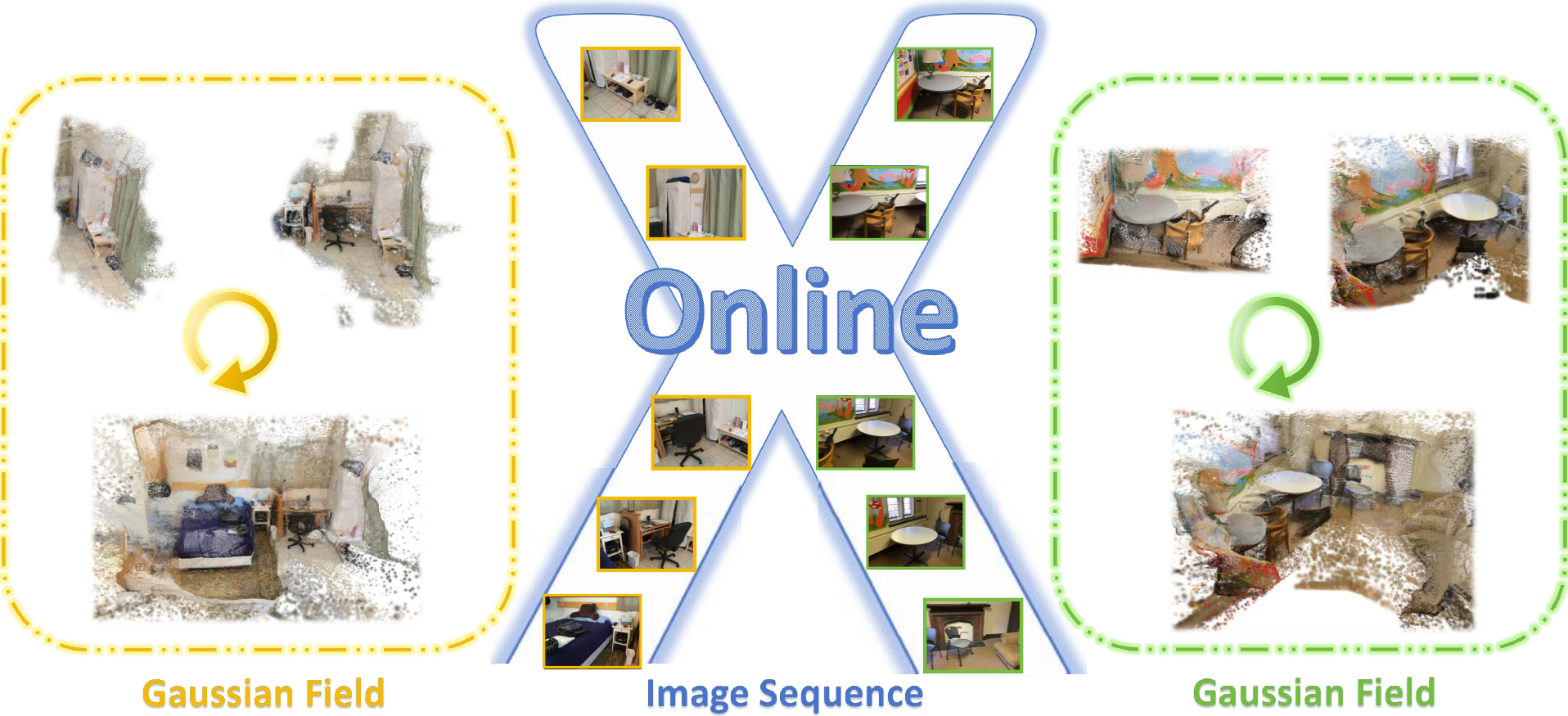}
    \caption{We introduce \textbf{OnlineX}, a framework for continuous and progressive 3D scene reconstruction from streaming images. Our core contribution is a active-to-stable state evolution paradigm, which effectively mitigates long-term drift by decoupling the processing of high-fidelity active local details from the maintenance of a stable global structure.}
    \label{fig:teaser}
    \vspace{-12pt}
  \end{figure*}

In this paper, we propose OnlineX, a generalizable model for online 3D Gaussian reconstruction and understanding built upon a Active-to-Stable state evolution, as shown in Figure \ref{fig:teaser}. First, we perform a pairwise interaction between the current and preceding frames to extract the active state representing the relative information of per-pixel geometry and appearance. Then, the relative active feature is integrated with our stable global anchor state to compute an updated and globally consistent pose feature for the current frame. Finally, the updated global pose feature implicitly projects the previously extracted relative geometry into a globally consistent structure, which avoids the potential instability of explicit pose transformation. In this way, we decouple local active state extraction from stable state maintenance, which not only alleviates the representational bottleneck of the global state, but also provides a more informative signal for its update.

Motivated by the coherent distributions of appearance and semantics across multiple views, our framework jointly models both visual appearance and language fields within the unified online paradigm. Furthermore, we introduce an implicit Gaussian fusion module that merges duplicate overlapped Gaussian primitives and integrates their features, which are then decoded into the final Gaussian primitives. Finally, we employ an auxiliary supervision strategy to ensure the stable convergence of the entire online framework. Extensive experiments on multiple datasets demonstrate that OnlineX consistently achieves superior performance in novel view synthesis and open-vocabulary semantic segmentation across input view sequences of varying lengths, while enabling real-time inference.
\section{Related Work}

\paragraph{Generalizable 3D Reconstruction.} 
3D Gaussian Splatting (3DGS)~\cite{kerbl20233d, yu2024mip, ren2024octree, luiten2024dynamic} has recently emerged as a promising alternative for real-time 3D scene reconstruction. While early 3DGS methods require per-scene optimization, generalizable feed-forward models~\cite{chen2024mvsplat, charatan2024pixelsplat, zou2024triplane, zheng2024gps, wewer2024latentsplat, li2024ggrt} were introduced to eliminate this need but remain dependent on pre-computed camera poses from offline SfM tools like COLMAP~\cite{schonberger2016structure}. To overcome this limitation, recent pose-free approaches~\cite{wang2024dust3r, leroy2024grounding, smart2024splatt3r, ye2024no, zhang2025flare} jointly estimate poses and reconstruct scenes. Nonetheless, these methods typically operate on fixed-size sets of images, rather than continuous, online reconstruction. In contrast, our work targets generalizable online 3D reconstruction from streaming RGB input towards real-time applications.

\paragraph{3D Scene Understanding.} 
The integration of 3DGS with vision foundation models like SAM~\cite{kirillov2023segment} and CLIP~\cite{radford2021learning} has recently gained momentum for enabling open-world 3D scene understanding~\cite{qin2024langsplat,ye2024gaussian,xu2024tiger,qu2024goi,ji2025fastlgs,hu2024semantic,qiu2024feature}. Methods like LangSplat~\cite{qin2024langsplat} and Gaussian Grouping~\cite{ye2024gaussian} attach semantic features to Gaussians but are limited by their reliance on per-scene optimization. While some recent feed-forward networks~\cite{li2025scenesplat} offer generalizability, they treat reconstruction and understanding as separate, post-hoc tasks. In contrast, our work introduces a unified, end-to-end framework that learns both visual appearance and language fields concurrently from source images, eliminating the need for per-scene optimization or specialized network modules.

\paragraph{3D Online Paradigm.}
The paradigm of building and interpreting 3D scenes from sequential input is a long-standing goal in computer vision~\cite{engel2014lsd,forster2016svo,teed2021droid,zhu2024nicer,sun2021neuralrecon,zhang2022nerfusion,wang20243d,wang2025continuous,choy20163d,kar2017learning, chen2025long3r}. However, modern learning-based methods often face a fundamental architectural trade-off. Models employing explicit spatial memory like Spann3R~\cite{wang20243d} and LONG3R~\cite{chen2025long3r} incur unsustainable memory overhead, whereas those using a compact, implicit state like CUT3R~\cite{wang2025continuous} are susceptible to geometric drift. Similar trade-offs exist in parallel online perception research~\cite{liu2022ins,huang2021supervoxel,xu2024memory,zhang2020fusion,xia2025scenepainter,wu2023anyview}. Our framework addresses these challenges by introducing a new online paradigm, which leverages the efficiency of implicit representations but resolves their inherent bottleneck by processing active and stable state in two decoupled streams that are then cohesively fused to achieve both high-fidelity detail and long-term consistency.


\begin{figure*}[t]
    \centering
    \includegraphics[width=1.0\linewidth]{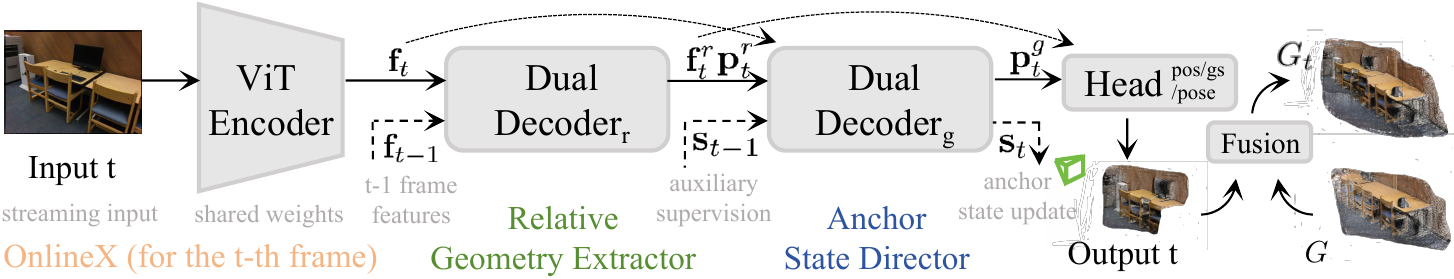}
    \caption{\textbf{Overall architecture of OnlineX.} Our framework features a two-stage, active-to-stable pipeline. First, the Relative Geometry Extractor processes consecutive frames to capture high-fidelity active relative information. The Anchor State Director then uses this local information to recurrently update its stable global state, yielding a globally consistent representation for the final output. The diagram illustrates this process for a single time step, which would be sequentially repeated for each frame in the input stream. Dashed lines represent information passed from the previous time step or carried over to the next.}
    \label{fig:pipeline}
    \vspace{-12pt}
\end{figure*}

\section{Method}
\label{method}
In this section, we present the overall framework of OnlineX, as illustrated in Figure \ref{fig:pipeline}. We first define the problem formulation of online 3D Gaussian Splatting reconstruction with understanding in Section \ref{sec:problem_formulation}. We further introduce the active-to-stable state evolution paradigm for online 3D Gaussian reconstruction in Section \ref{sec:relative_aggregation} and Section \ref{sec:global_projection}. We then propose the implicit Gaussian fusion module to merge overlapped 3D Gaussians across multiple viewpoints in Section \ref{sec:adaptive_fusion}. Finally, we detail the training objectives and strategies for the whole framework in Section \ref{sec:training_details}.

\subsection{Problem Formulation}
\label{sec:problem_formulation}
\paragraph{Gaussian Primitive Representation.} Given a streaming sequence of pose-free RGB frames $\left\{I_t\right\}_{t=1}^{T}$ as input and for the $t$-th frame, our model is to concurrently predict the corresponding 3D Gaussian representation $G_t$, defined as:
\begin{equation}
    G_t = \left\{ \left( \mathbf{\mu}_t^i, \mathbf{r}_t^i, \mathbf{s}_t^i, \mathbf{\alpha}_t^i, \mathbf{c}_t^i, \mathbf{l}_t^i \right) \right\}_{i=1}^{N_t},
\end{equation}where $N_t$ is the number of Gaussians associated with the $t$-th frame. Each Gaussian encapsulates the basic components of the vanilla 3DGS formulation~\cite{kerbl20233d}, including the center position $\mathbf{\mu}_t^i \in \mathbb{R}^3$, rotation quaternion $\mathbf{r}_t^i \in \mathbb{R}^4$, scale factor $\mathbf{s}_t^i \in \mathbb{R}^3$, opacity $\mathbf{\alpha}_t^i \in \mathbb{R}$ and color information $\mathbf{c}_t^i \in \mathbb{R}^3$ expressed using spherical harmonics. 

In addition, we add the language feature $\mathbf{l}_t^i \in \mathbb{R}^K$ for each Gaussian to reconstruct the language field with geometry and appearance as an exploratory extension. Inspired by LangSplat~\cite{qin2024langsplat}, to reduce the memory and computational costs, we regress the semantic feature at a low dimensional space, with the feature dimension $K \ll D$, where $D = 512$ denotes the original dimension of the 2D semantic features extracted from CLIP~\cite{radford2021learning} and we set $K$ as 16 by default.

\paragraph{Rendering Process.} The set of 3D Gaussians $G$ is subsequently employed for rendering novel-view RGB images and language maps as the following alpha blending process:
\begin{equation}
    \left\{
    \begin{aligned}
    \mathrm{C}(v) &= \sum_{i\in N} \mathbf{c}_i \mathbf{\alpha}_i \prod_{j=1}^{i-1} (1-\mathbf{\alpha}_j), \\
    \mathrm{L}(v) &= \sum_{i\in N} \mathbf{l}_i \mathbf{\alpha}_i \prod_{j=1}^{i-1} (1-\mathbf{\alpha}_j),
    \end{aligned}
    \right.
\end{equation}where $\mathrm{C}(v)$ and $\mathrm{F}(v)$ represent the rendered color and language feature at pixel $v$ in the novel view image, and $N$ is the number of Gaussians that the ray passes through.

\paragraph{Generalizable Online Reconstruction.}
Conventional feed-forward 3DGS methods typically process the entire frame sequence as input to infer the complete set of Gaussians as follows:
\begin{equation}
    f(\{I_t\}_{t=1}^{T};\mathbf{\theta}) = G,
\end{equation}where $f$ represents the feed-forward network and $\theta$ denotes its learnable parameters. While straightforward, this design necessitates complete video sequences, which hinders its capacity for continuous 3D reconstruction. To address these limitations, we propose an incremental 3DGS reconstruction framework capable of incessantly regressing Gaussians for each incoming frame, without reliance on the whole preceding frames, which can be formulated as:
\begin{align}
    f(I_t, \mathbf{h}_{t-1}; \mathbf{\theta}) =  G_t, \mathbf{h}_t, \quad t = 1, \dots, T,
\end{align}where $\mathbf{h}$ denotes the historical information from preceding frames that would be recurrently updated with new frame input. The output $G_t$ is then incrementally integrated into the global 3DGS representation $G$ as the accumulated reconstruction result of the observed environment.

\subsection{Relative Geometry Extractor}

\label{sec:relative_aggregation}
In this section, we introduce the Relative Geometry Extractor stage of our framework, which is primarily responsible for regressing detailed relative geometry and Gaussian parameters of the current frame based on the preceding frame. This stage not only alleviates the representational burden on the global anchor state for storing high-frequency active details, but also distills the dense local information into an informative and structured signal that effectively guides the subsequent stable anchor state modeling stage.

\paragraph{Encoder and Decoder.}
The RGB image of the each frame are first patchified and flattened into sequences of image tokens, and then fed into a ViT~\cite{dosovitskiy2020image} encoder separately. The encoder shares the same weights for different views. For the $t$-th frame, the per-pixel extracted features of the current frame $\mathbf{f}_t$ and the preceding frame $\mathbf{f}_{t-1}$ are each concatenated with a learnable pose token, which serves as a learnable pose embedding to regress the relative pose information based on the preceding frame. The concatenated features are then fed into a dual ViT decoder module~\cite{wang2024dust3r, leroy2024grounding}, where hidden features from each view interact with the other view through cross-attention layers in each attention block, facilitating relative information extraction. Finally, the output features $\mathbf{p}^r_t$, $\mathbf{f}^r_t$ and $\mathbf{f}^r_{t-1}$ are processed by the following prediction heads for proper supervision and direction, while the $\mathbf{p}^r_t$ and $\mathbf{f}^r_t$ serve as input to the recurrent modeling in the global anchor state stage. Specially, for the first frame, we obtain output features $\mathbf{f}^r_1$ when processing the second frame. The whole procedure could be formulated as:

\begin{align}
    \mathbf{f}_t &= \text{Encoder}(I_t), \\
     [\mathbf{p}^{r}_t,\mathbf{f}^{r}_t], [\mathbf{p}'_0,\mathbf{f}^{r}_{t-1}]&= \text{Decoder}_r([\mathbf{p}_1, \mathbf{f}_t],[\mathbf{p}_0,\mathbf{f}_{t-1}]).
\end{align}

\paragraph{Relative Prediction Heads.} 
The output features $\mathbf{p}^r_t$, $\mathbf{f}^r_t$ and $\mathbf{f}^r_{t-1}$ encapsulate the relative geometry, appearance and pose information between the current frame and the last frame. These features are then processed by three distinct prediction heads to regress the following relative outputs:
\begin{align}
X_t^r, C_t^r &= \text{Head}^{\text{pos}}_r(\mathbf{f}^r_t, \mathbf{f}^r_{t-1}), \\
G_t^r &= \text{Head}^{\text{gs}}_r(\mathbf{f}_t^r, \mathbf{f}_{t-1}^r), \\
P_t^r &= \text{Head}^{\text{pose}}_r(\mathbf{p}_t^r).
\end{align}
Specifically, $X_t^r$ and $C_t^r$ are the predicted per-pixel Gaussian centers and their corresponding confidence maps; $G_t^r$ encapsulates all other Gaussian attributes such as color, scale, rotation, language features and opacity; and $P^r_t$ is the estimated relative camera pose. The $\text{Head}^{\text{pos}}_r$ and $\text{Head}^{\text{gs}}_r$ follow a DPT~\cite{ranftl2021vision} architecture, and $\text{Head}^{\text{pose}}_r$ is a simple MLP network. Note that these Gaussian outputs are a joint prediction of both the current and preceding frames based on the preceding frame's coordinate system. Although these relative outputs do not directly constitute the final global reconstruction, they provide a crucial auxiliary supervision signal. This intermediate supervision is essential for stabilizing the end-to-end training of our online framework.

\subsection{Anchor State Director} 
\label{sec:global_projection}
This section details the Anchor State Director stage of our framework, which is responsible for generating the final, globally consistent representation. The process begins by introducing the stable Anchor State which stores the historical global context. We further extract a compact feature vector from the per-pixel features and the relative pose features of the current frame. This vector is then updated with the Anchor State through a recurrent update mechanism. Finally, this updated feature vector, which now encapsulates the current global structure, is fused with the high-frequency local details from the preceding stage to produce the globally-aware representation for the current frame.

\paragraph{Recurrent Modeling.}
The Anchor State is our stable memory that encapsulates the accumulated global structure of the scene up to the current frame. At the beginning of a sequence, the initial state $s_0$ is instantiated from a set of learnable tokens, which are trained to encode a prior of generic 3D scene structures. For each subsequent frame, the Anchor State is iteratively computed from the previous state and represents the extended global structure. By offloading the responsibility for processing high-frequency, per-frame details to the relative stage, our design prevents the Anchor State from undergoing volatile updates and heavy memory overhead, thereby preserving its integrity as a stable repository for the scene's global structure.

To integrate information from the current frame into the global Anchor State, we construct a compact feature vector for the current frame by concatenating three components: the relative pose features $\mathbf{p}^r_t$, the globally-pooled features from the relative stage $\mathbf{\bar{f}}^r_t$ and the globally-pooled features from the initial encoder $\mathbf{\bar{f}}_t$. The two components—the compact feature vector and the Anchor State $\mathbf{s}_{t-1}$—are then jointly fed into a pair of interconnected transformer decoders. This recurrent update is formulated as:
\begin{equation}
 \mathbf{p}^{g}_t, \mathbf{s}_t= \text{Decoder}_{g}([\mathbf{\bar{f}}_t, \mathbf{\bar{f}}^r_t, \mathbf{p}^r_t], \mathbf{s}_{t-1}).
\end{equation}
Here, $\mathbf{s}_{t-1}$ and $\mathbf{s}_t$ denote the Anchor State tokens before and after the interaction, respectively, while $\mathbf{p}^{g}_t$ represents the resulting global pose feature based on the first frame. This process is bidirectional: the input feature vector queries the historical context within $\mathbf{s}_{t-1}$ to produce the global pose feature $\mathbf{p}^g_t$. Concurrently, the Anchor State incorporates information from the current frame's features, yielding the updated state $\mathbf{s}_{t}$ to be passed to the subsequent time step. 

\paragraph{Global Prediction Heads.}
The global prediction heads receive two primary inputs: the relative per-pixel features $\mathbf{f}_t^r$ from the preceding stage, which encapsulate high-fidelity local geometry, and the updated global pose feature $\mathbf{p}_t^g$, which provides the global context. These features are then processed by a set of distinct heads to regress the final global outputs based on the first frame's coordinate system:
\begin{align}
    X_t^g, C_t^g &= \text{Head}_g^{\text{pos}}(\mathbf{f}_t^r, \mathbf{p}_t^g), \\
    G_t^g &= \text{Head}_g^{\text{gs}}(\mathbf{f}_t^r, \mathbf{p}_t^g), \\
    P_t^g &= \text{Head}_g^{\text{pose}}(\mathbf{p}_t^g).
\end{align}
Similar to the relative stage, these global outputs are produced by three distinct prediction heads. The DPT-based $\text{Head}_g^{\text{pos}}$ regresses the final Gaussian centers $X_t^g$ and their corresponding confidence maps $C_t^g$. Concurrently, the DPT-based $\text{Head}_g^{\text{gs}}$ predicts all other Gaussian attributes $G_t^g$, including language features, while the MLP-based $\text{Head}_g^{\text{pose}}$ outputs the definitive global pose $P_t^g$. Crucially, within the DPT-based heads, we perform cross-attention between the local geometric features $\mathbf{f}_t^r$ and the global pose feature $\mathbf{p}_t^g$. This mechanism performs an implicit transformation, conditioning the local geometry with the global context directly in the feature space. This learned, feature-based alignment is more flexible and robust compared to applying a rigid, explicit pose transformation. Thus, the outputs of this stage ($X_t^g, G_t^g$) represent a sophisticated fusion of high-fidelity local geometry and consistent global structure.

\subsection{Implicit Gaussian Fusion}
\label{sec:adaptive_fusion}
To address the issue of redundant Gaussians in prior 3DGS methods, which often rely on simplistic, opacity-based pruning, we introduce our Implicit Gaussian Fusion module. Inspired by~\cite{gao2023surfelnerf, sun2021neuralrecon}, this module resolves this by adaptively identifying and merging nearby primitives in the latent space. For each new Gaussian $g_t$ (with center $\mathbf{x}_t$ and confidence $c_t$), we first identify its neighborhood $\mathcal{N}_t$ by finding all existing Gaussians within the same spatial voxel. The fusion process then refines both the geometric position and the latent attributes. The new center $\mathbf{x}'_t$ is computed as a confidence-weighted average of the neighborhood:
\begin{equation}
    \mathbf{x}'_t = \frac{c_t \mathbf{x}_t + \sum_{i \in \mathcal{N}_t} c_i \mathbf{x}_i}{c_t + \sum_{i \in \mathcal{N}_t} c_i},
    \label{eq:center_fusion}
\end{equation}
while the latent feature $\mathbf{g}_t$ is updated by fusing it with the weighted-average of its neighboring features $\tilde{\mathbf{g}}_n$ using a small MLP network:
\begin{align}
    \tilde{\mathbf{g}}_n &= \frac{\sum_{i \in \mathcal{N}_t} c_i \mathbf{g}_i}{\sum_{i \in \mathcal{N}_t} c_i},\\
    \mathbf{g}'_t &= \text{MLP}([\mathbf{g}_t, \tilde{\mathbf{g}}_n]).
\end{align}
This iterative, latent-space refinement process yields a more compact and globally consistent scene representation.

\subsection{Training Details}
\label{sec:training_details}

\paragraph{Training Objectives.}
Our framework is trained end-to-end using a composite loss function that includes terms for pose $\mathcal{L}_{\text{pose}}$, visual rendering $\mathcal{L}_{\text{render}}$, and language rendering $\mathcal{L}_{\text{lang}}$. Crucially, we employ a auxiliary supervision strategy where these losses are applied at both the intermediate relative stage and the final global stage of our architecture. This intermediate objective ensures that the network first learns to extract high-fidelity local representations, which provides a stable foundation for the subsequent global update. The total loss is a weighted sum of the primary global and auxiliary relative losses:
\begin{align}
    \mathcal{L}_{\text{total}} &= \mathcal{L}_{\text{global}} + \lambda_{\text{aux}} \mathcal{L}_{\text{relative}}, \\
    \text{where } \mathcal{L}_{(\cdot)} &= \lambda_1 \mathcal{L}_{\text{pose}} + \lambda_2 \mathcal{L}_{\text{render}} + \lambda_3 \mathcal{L}_{\text{lang}}.
\end{align}
We adopt L2 loss for $\mathcal{L}_{\text{pose}}$, a combination of MSE and LPIPS~\cite{zhang2018unreasonable} loss for $\mathcal{L}_{\text{render}}$ and negative cosine similarity for $\mathcal{L}_{\text{lang}}$, which are detailed in the supplementary material.

\begin{table*}[tp]
	\centering
	\setlength{\abovedisplayskip}{0pt}
	\setlength{\belowdisplayskip}{0pt}
    \small
	\caption{\textbf{Quantitative comparison of novel view synthesis on RE10K~\cite{zhou2018stereo}.} Our method achieves comparable performance with previous SOTA methods on few-view settings, and outperforms them with more views input. }
    \vspace{2pt}
    \scalebox{0.9}{
	\begin{tabular}{l|w{c}{1.0cm}w{c}{1.0cm}w{c}{1.1cm}w{c}{1.0cm}w{c}{1.0cm}w{c}{1.1cm}w{c}{1.0cm}w{c}{1.0cm}w{c}{1.1cm}}
		\toprule
		 \multirow{2}{*}{Method} & \multicolumn{3}{c}{2 views} & \multicolumn{3}{c}{4 views} & \multicolumn{3}{c}{8 views}  \\
          & PSNR $\uparrow$ & SSIM $\uparrow$ & LPIPS $\downarrow$ & PSNR $\uparrow$ & SSIM $\uparrow$ & LPIPS $\downarrow$ & PSNR $\uparrow$ & SSIM $\uparrow$ & LPIPS $\downarrow$ \\
		\midrule

        MVSplat~\cite{chen2024mvsplat}&24.73&0.829&0.171&21.91&0.753&0.358&20.41&0.696&0.434\\
        NoPoSplat~\cite{ye2024no}&\underline{25.59}&\textbf{0.849}&\textbf{0.140}&22.63&0.771&0.302&21.08&0.671&0.412 \\
        FLARE~\cite{zhang2025flare}&24.83&0.834&0.165&\underline{23.16}&\underline{0.794}&\underline{0.218}&22.41&0.765&0.314 \\
        \midrule
        Spann3R~\cite{wang20243d}+GS&22.43&0.761&0.317&22.07&0.732&0.374&21.86&0.701&0.396 \\
        CUT3R~\cite{wang2025continuous}+GS&23.12&0.798&0.208&22.91&0.775&0.241&\underline{22.78}&\underline{0.774}&\underline{0.236} \\
        \textbf{Ours}&\textbf{25.78}&\underline{0.845}&\underline{0.144}&\textbf{25.56}&\textbf{0.839}&\textbf{0.147}&\textbf{25.59}&\textbf{0.841}&\textbf{0.148} \\

		\bottomrule
	\end{tabular}}
    \label{tab:re10k}
\end{table*}

\begin{table*}[tp]
	\centering
	\setlength{\abovedisplayskip}{0pt}
	\setlength{\belowdisplayskip}{0pt}
    \small
	\caption{\textbf{Quantitative comparison of novel view synthesis on ScanNet~\cite{dai2017scannet}.} Our method outperforms previous SOTA methods on all input view settings, even larger gains on 30-view setting.}
    \vspace{2pt}
    \scalebox{0.9}{
	\begin{tabular}{l|w{c}{1.0cm}w{c}{1.0cm}w{c}{1.1cm}w{c}{1.0cm}w{c}{1.0cm}w{c}{1.1cm}w{c}{1.0cm}w{c}{1.0cm}w{c}{1.1cm}}
		\toprule
		 \multirow{2}{*}{Method} & \multicolumn{3}{c}{10 views} & \multicolumn{3}{c}{20 views} & \multicolumn{3}{c}{30 views}  \\
          & PSNR $\uparrow$ & SSIM $\uparrow$ & LPIPS $\downarrow$ & PSNR $\uparrow$ & SSIM $\uparrow$ & LPIPS $\downarrow$ & PSNR $\uparrow$ & SSIM $\uparrow$ & LPIPS $\downarrow$ \\
		\midrule
                
        MVSplat~\cite{chen2024mvsplat}&20.03&0.686&0.414&18.72&0.561&0.511&16.02&0.505&0.591\\
        NoPoSplat~\cite{ye2024no}&21.34&0.729&0.360&19.42&0.635&0.469&18.73&0.574&0.502 \\
        FLARE~\cite{zhang2025flare}&21.75&0.751&0.330&19.91&0.652&0.447&19.15&0.619&0.485 \\
        \midrule
        Spann3R~\cite{wang20243d}+GS&21.54&0.732&0.357&\underline{20.91}&\underline{0.712}&\underline{0.367}&\underline{20.42}&\underline{0.689}&\underline{0.387} \\
        CUT3R~\cite{wang2025continuous}+GS&\underline{21.97}&\underline{0.762}&\underline{0.323}&20.79&0.709&0.382&20.01&0.687&0.401 \\
        \textbf{Ours}&\textbf{24.13}&\textbf{0.811}&\textbf{0.174}&\textbf{23.86}&\textbf{0.782}&\textbf{0.189}&\textbf{23.73}&\textbf{0.769}&\textbf{0.194} \\
		\bottomrule
	\end{tabular}}
    \label{tab:scannet}
\end{table*}

\section{Experiments}
\subsection{Experimental Settings}
\label{sec:exp}
\paragraph{Datasets.} We train and evaluate our model on two widely-used real-world datasets: RealEstate10k (RE10K)~\cite{zhou2018stereo} and ScanNet~\cite{dai2017scannet}. As our primary benchmarks, RE10K is used for performance evaluation on video sequences with a limited spatial scope, while ScanNet serves as the basis for our room-scale reconstruction task. To evaluate the generalization ability of our model, we further perform zero-shot evaluation on the DL3DV dataset~\cite{ling2024dl3dv}.

\paragraph{Baselines.} To assess the effectiveness of our OnlineX framework, we compare it against several state-of-the-art feed-forward 3D reconstruction methods, which fall into two main categories: (1) offline feed-forward 3DGS approaches, including MVSplat~\cite{chen2024mvsplat}, NoPoSplat~\cite{ye2024no} and FLARE~\cite{zhang2025flare}; and (2) online feed-forward pointmap prediction methods, such as Spann3R~\cite{wang20243d} and CUT3R~\cite{wang2025continuous}. For the offline 3DGS approaches, we provide all input views simultaneously, allowing them to perform an offline reconstruction. This setting is inherently less challenging than our online scenario, which must build the scene progressively from a sequential stream without access to future frames. For the online pointmap methods, since they do not natively produce Gaussian representations, we adapt them for a fair comparison by extending their architectures with a Gaussian Splatting prediction head identical to our own and subsequently fine-tuning them on the corresponding datasets. In addition, for the 3D scene understanding task, we compare our method to the per-scene optimization-based method LangSplat~\cite{qin2024langsplat} and Gaussian-Grouping (GS-Group)~\cite{ye2024gaussian} with the same input views.

\paragraph{Implementation Details.} During training, we sample sequences of randomly varying lengths (from 4 to 15 views) to teach the model the principles of online, iterative modeling, thereby ensuring its ability to generalize to longer, unseen sequences at inference time. he sampling interval between neighboring frames is set to 10 for RE10K and 20 for ScanNet, resulting in a moderate view overlap. We train our OnlineX model using the AdamW optimizer~\cite{loshchilov2017fixing} with an initial learning rate of $5 \times 10^{-5}$ for a total of 30,000 iterations on 4 NVIDIA RTX A6000 GPUs with an effective batch size of 8. The loss weights $\lambda_{\text{aux}}$, $\lambda_{1}$, $\lambda_{2}$ and $\lambda_{3}$ are set to 0.8, 1, 1 and 0.5, respectively.  We adopt pre-trained weights of MASt3R~\cite{leroy2024grounding} with adaptive adjustment.



\begin{figure*}[t]
    \centering
    \includegraphics[width=1.0\linewidth]{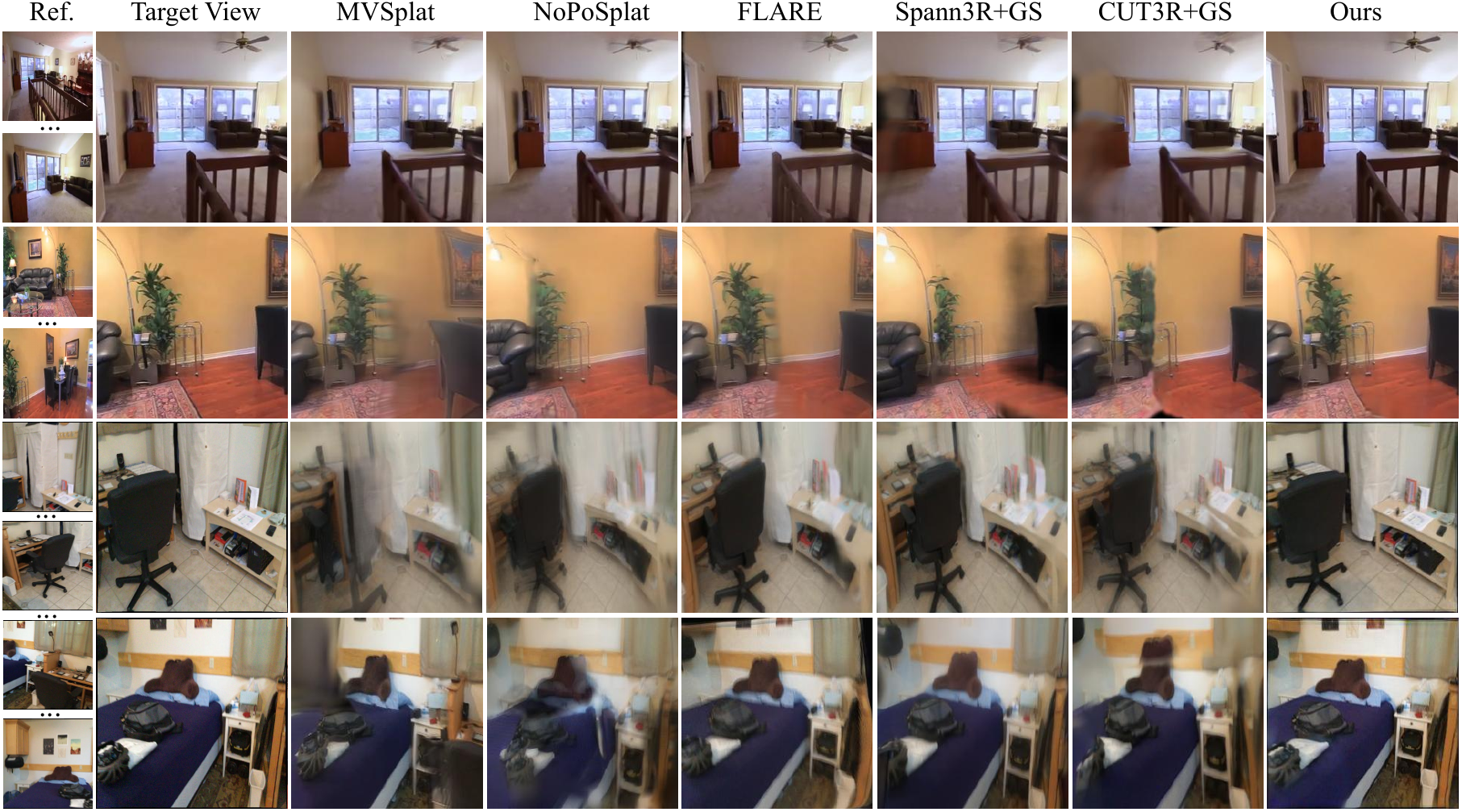}
    \caption{\textbf{Qualitative comparison for novel view synthesis on RE10K (top two rows) and ScanNet (bottom two rows).} We adopt the 4-view setting for RE10K and 15-view setting for ScanNet. }
    \label{fig:vis}
\end{figure*}

\begin{table}[t]
	\centering
	\small
	\caption{\textbf{Camera pose estimation on ScanNet.} We report Absolute Translation Error (ATE), Relative Translation Error (RPE trans), and Relative Rotation Error (RPE rot).}
	\label{tab:pose_scannet}
	\begin{tabular}{l ccc}
		\toprule
		Method & ATE$\downarrow$ & RPE trans$\downarrow$ & RPE rot $\downarrow$ \\
		\midrule
		Spann3R~\cite{wang20243d} & \underline{0.096} & 0.023 & 0.661 \\
		CUT3R~\cite{wang2025continuous} & 0.099 & \underline{0.022} & \underline{0.600} \\
		\textbf{Ours} & \textbf{0.085} & \textbf{0.019} & \textbf{0.550} \\
		\bottomrule
	\end{tabular}
\end{table}
\begin{table}[t]
	\centering
	\setlength{\abovedisplayskip}{0pt}
	\setlength{\belowdisplayskip}{0pt}
    \small
	\caption{\textbf{Quantitative comparison of semantic segmentation on ScanNet~\cite{dai2017scannet}.} We report the average IoU scores ($\%$) and average accuracy ($\%$). We denote Gaussian-Grouping~\cite{ye2024gaussian} as GS-Group.}
    \vspace{2pt}
    \scalebox{0.9}{
    \begin{tabular}{l|cccc}
		\toprule
		 \multirow{2}{*}{Method} & \multicolumn{2}{c}{5 views} & \multicolumn{2}{c}{15 views}  \\
          & mIoU $\uparrow$ & mAcc $\uparrow$ & mIoU $\uparrow$ & mAcc $\uparrow$ \\
		\midrule
        
        LangSplat~\cite{qin2024langsplat}&\underline{54.63}&\underline{71.15}&\underline{56.91}&\underline{73.41}\\
        GS-Group~\cite{ye2024gaussian}&51.61&68.32&53.26&67.63  \\
        \textbf{Ours}&\textbf{58.83}&\textbf{77.12}&\textbf{57.79}&\textbf{76.36} \\
		\bottomrule
	\end{tabular}}
    \label{tab:scannet_lang}
    \vspace{-2pt}
\end{table}

\subsection{Results}
\paragraph{Novel View Synthesis.}
Table \ref{tab:re10k} and Table \ref{tab:scannet} present the novel view synthesis (NVS) results of our proposed OnlineX on the RE10K~\cite{zhou2018stereo} and ScanNet~\cite{dai2017scannet} datasets, respectively. For a comprehensive evaluation, we test on varying sequence lengths: 2, 4, and 8 views for RE10K, and 10, 20, and 30 views for ScanNet. In few-view settings, our method achieves performance comparable to offline reconstruction baselines. As the number of views increases, OnlineX demonstrates stable performance and consistently outperforms competing online approaches. These results demonstrate our model's strong performance in both sparse-view and long-term online scenarios.

\begin{figure*}[t]
    \centering
    \includegraphics[width=1.0\linewidth]{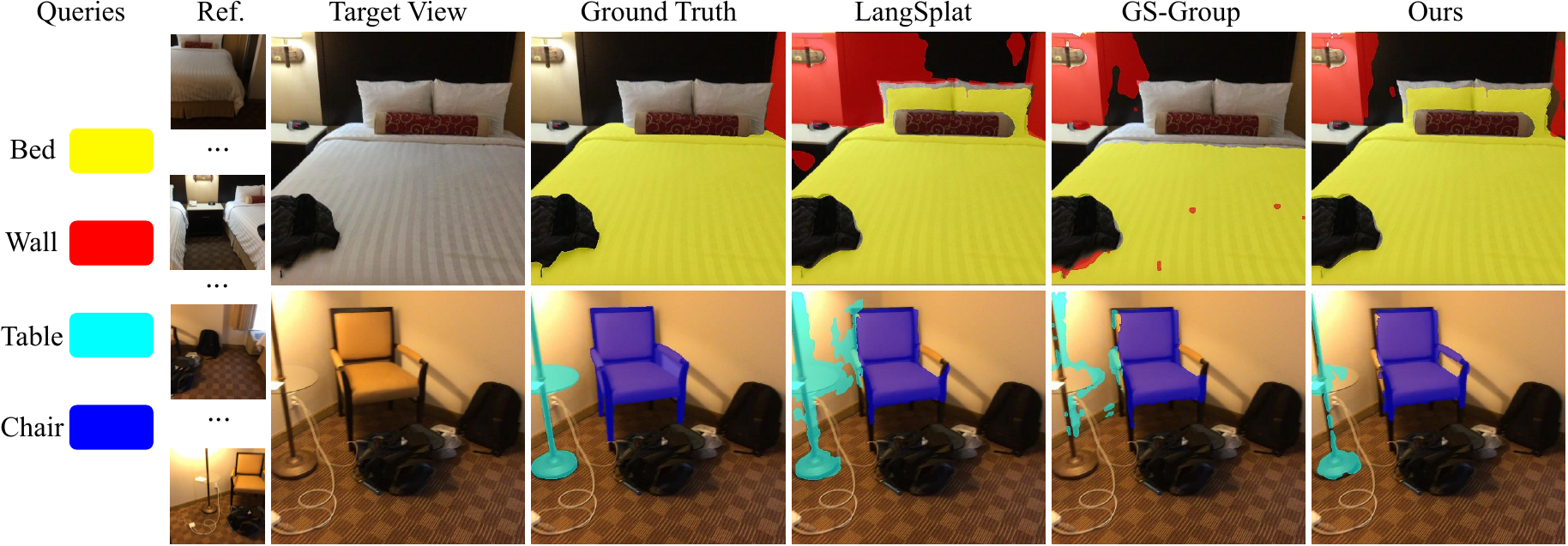}
    \caption{\textbf{Qualitative comparison for semantic segmentation on ScanNet. } Here we showcase one scene with 15 input views. The masks predicted by ours contain more complete regions than other methods, such as the "Wall" prompt, which also surpasses the GT masks.}
    \label{fig:vis_lang}
    \vspace{3pt}
\end{figure*}

\begin{figure}[t]
    \centering
    \includegraphics[width=1.0\linewidth]{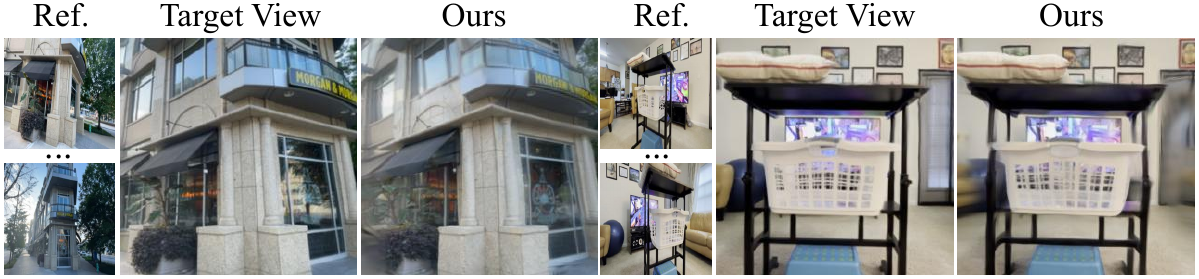}
    \caption{\textbf{Qualitative results for zero-shot generalization on DL3DV. } Our model can easily transfer to out-of-distribution data.}
    \label{fig:ood}
    \vspace{-3pt}
\end{figure}

Figure \ref{fig:vis} further shows the visualization results of NVS compared with baseline methods. It can be observed that our method significantly surpasses the baselines in predicting accurate global geometry, capturing fine-grained details, and reducing artifacts caused by overlapped Gaussians. 

\paragraph{Camera Pose Estimation.}
We evaluate the camera pose estimation accuracy of our method on the ScanNet dataset~\cite{dai2017scannet} with 30 input views. Following standard protocols~\cite{chen2024leap, zhao2022particlesfm, zhang2024monst3r}, we report Absolute Translation Error (ATE), Relative Translation Error (RPE trans), and Relative Rotation Error (RPE rot) after Sim(3) alignment with the ground truth. We compare our approach against online methods that similarly do not require camera calibration, specifically CUT3R~\cite{wang2025continuous} and Spann3R~\cite{wang20243d}. As shown in Table~\ref{tab:pose_scannet}, our OnlineX framework consistently outperforms both baselines across all three metrics, demonstrating the effectiveness of our decoupled architecture in maintaining a more accurate and robust camera trajectory.

\paragraph{Open-Vocabulary Semantic Segmentation.} For open-vocabulary segmentation, we query the rendered 2D language feature map by computing the per-pixel cosine similarity between its features and a given text embedding. This process yields a confidence map, where regions with high similarity scores are taken as the final segmentation result for the queried object. We evaluate our method using mIoU and mAcc on the ScanNet datasets, as shown in Table \ref{tab:scannet_lang}. Compared with existing 3D language field methods such as LangSplat~\cite{qin2024langsplat} and Gaussian Grouping~\cite{ye2024gaussian}, our approach achieves superior performance across both metrics. Visualization results in Figure \ref{fig:vis_lang} further show that our method accurately segments objects with detailed boundaries, highlighting the strength of our unified visual and semantic understanding in capturing visual guidance.

\paragraph{Cross-Dataset Generalization.} We also assess the zero-shot generalization ability of our model, where we train on RE10K~\cite{zhou2018stereo} and directly evaluate it on DL3DV~\cite{ling2024dl3dv} for NVS task with 6 input views. As shown in Figure \ref{fig:ood}, our method in general shows superior zero-shot performance on out-of-distribution data. This improved generalization is largely attributed to the unified online architecture in our method, which enhances its adaptability across diverse scene types and varying lengths of input sequences.


\paragraph{Runtime Analysis.}
Our method achieves 23 frames per second (FPS) on 256×256 inputs using a single NVIDIA RTX A6000 GPU, supporting real-time applications. The inference time and GPU memory usage are comparable to that of CUT3R~\cite{wang2025continuous} and are significantly faster and lower than Spann3R~\cite{wang20243d} as shown in Table \ref{tab:runtime}.

\begin{table}[t]
	\centering
    \small
	\caption{\textbf{Runtime and memory analysis.} We report the FPS and memory usage metric compared with other online methods.}
    \begin{tabular}{l|ccc}
		\toprule
		  Method & Spann3R~\cite{wang20243d} & CUT3R~\cite{wang2025continuous} & \textbf{Ours} \\
        \midrule
        FPS & 13.35  & \textbf{25.78} & \underline{23.12} \\
        Memory(GB)&32.73&\textbf{19.76}&\underline{21.64}\\
		\bottomrule
	\end{tabular}
    \label{tab:runtime}
\end{table}

\begin{table}[t]
    \centering
    \small
    \caption{\textbf{Quantitative results of the ablation study. } We report novel view synthesis metrics on ScanNet with 10 views.}
    \begin{tabular}{l|ccc}
		\toprule
		Method & PSNR $\uparrow$ & SSIM $\uparrow$ & LPIPS $\downarrow$\\
		\midrule
		w/o Relative Extractor & 20.85 & 0.712 & 0.368 \\
		w/o Anchor State & 19.92 & 0.685 & 0.441  \\
		w/o Implicit Transform & 16.52 & 0.533 & 0.552\\
		w/o Implicit GS Fusion & \underline{22.61} & \underline{0.778} & \underline{0.284} \\
		\midrule
		\textbf{Ours (Full Model)} & \textbf{24.13} & \textbf{0.811} & \textbf{0.174}  \\
      \bottomrule
    \end{tabular}
    \label{tab:ablation}
\end{table}

\subsection{Ablation Studies}

\begin{figure}[t]
    \centering
    \includegraphics[width=1.0\linewidth]{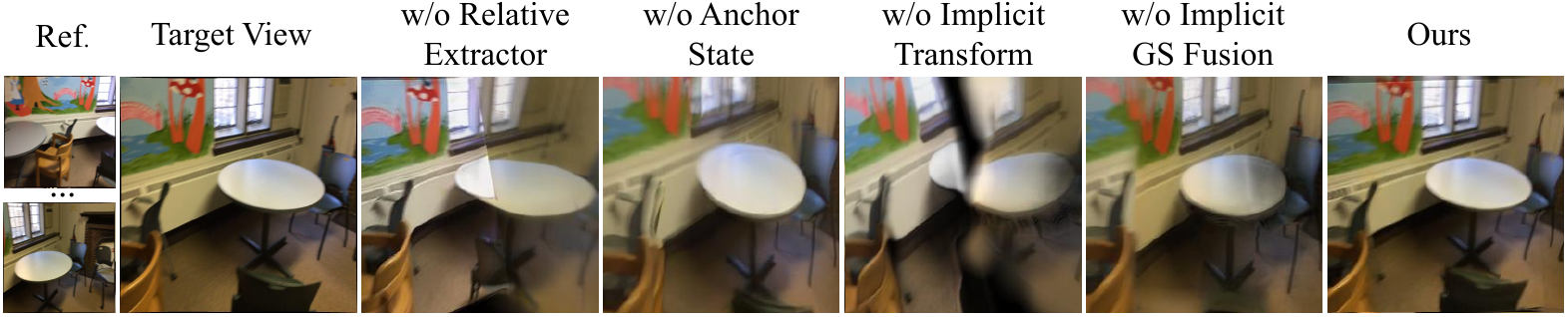}
    \caption{\textbf{Qualitative results of the ablation study.} From left to right, we visualize the results of our full model and four variants: without the Relative Geometry Extractor, without the Anchor State, without the Implicit Pose Transformation, and without the Implicit GS Fusion.}
    \label{fig:ablation}
    \vspace{-5pt}
\end{figure}

We conduct extensive ablation studies to verify the effectiveness of our key components (Table \ref{tab:ablation} and Figure \ref{fig:ablation}). Removing the Relative Geometry Extractor leads to a loss of fine-grained detail and inaccurate poses. Discarding the Anchor State results in severe camera drift, confirming its necessity for long-term consistency. Replacing our Implicit Pose Transformation with an explicit one causes visible seams between frames, demonstrating its importance for seamless reconstruction. Finally, omitting the Implicit GS Fusion module produces blurrier results with noticeable overlapping Gaussians at object boundaries.

\section{Conclusion}
In this paper, we presented OnlineX, a feed-forward framework for online 3D reconstruction and semantic understanding from only streaming RGB images. Our core contribution is an active-to-stable state evolution paradigm that resolves the inherent conflict between local fidelity and global consistency. By decoupling the extraction of active local geometry from the maintenance of a stable global state, OnlineX effectively mitigates the cumulative drift that challenges existing online methods. Furthermore, our unified framework jointly models visual appearance and language fields and incorporates an implicit Gaussian fusion module to ensure a compact and consistent representation. Extensive experiments validate that OnlineX achieves superior performance in both novel view synthesis and semantic understanding across varying sequence lengths, demonstrating a robust and scalable online 3D paradigm.
{
    \small
    \bibliographystyle{ieeenat_fullname}
    \bibliography{main}
}


\end{document}